\documentclass[twocolumn]{article}

\usepackage{balance}
\usepackage{cite}
\usepackage{graphicx}
\usepackage{amsmath}
\usepackage{amsfonts}
\usepackage{algorithm}
\usepackage{url}
\usepackage{subfig}

\usepackage{listings}
\usepackage[normalem]{ulem} % for results table
  \useunder{\uline}{\ul}{}
\usepackage{stmaryrd} % for \llbracket
\usepackage{mathtools} % for missing \rrightarrow
\newcommand{\rrightarrow}{\mathrel{\mathrlap{\rightarrow}\mkern1mu\rightarrow}}

\begin{document}

\title{Dependency Injection for\\ Programming by Optimization}

\author{Zoltan~A.~Kocsis$^1$  and Jerry~Swan$^2$\\
1. School of Mathematics, University of Manchester,\\
Oxford Road, Manchester M13 9PL, UK.\\
zoltan.kocsis@postgrad.manchester.ac.uk\\
2. Computer Science, University of York,\\
Deramore Lane, York, YO10 5GH, UK.
}

\maketitle

\begin{abstract}
Programming by Optimization tools perform automatic software configuration according to the specification supplied by a software developer. Developers specify design spaces for program components, and the onerous task of determining which configuration best suits a given use case is determined using automated analysis tools and optimization heuristics. However, in current approaches to Programming by Optimization, design space specification and exploration relies on external configuration algorithms, executable wrappers and fragile, preprocessed programming language extensions.

Here we show that the architectural pattern of Dependency Injection  provides a superior alternative to the traditional Programming by Optimization pipeline. We demonstrate that configuration tools based on Dependency Injection fit naturally into the software development process, while requiring less overhead than current wrapper-based mechanisms. Furthermore, the structural correspondence between Dependency Injection and context-free grammars yields a new class of evolutionary metaheuristics for automated algorithm configuration. We found that the new heuristics significantly outperform existing configuration algorithms on many problems of interest (in one case by two orders of magnitude). We anticipate that these developments will make Programming by Optimization immediately applicable to a large number of enterprise software projects.
\end{abstract}

%\noindent
%\textbf{Keywords}: Programming by Optimization, Search Based Software Engineering, Ant Programming, Dependency Injection, Combinatorial Optimization, Grammatical Evolution, Genetic Programming, Genetic Improvement.

\section{Introduction}

Proper configuration of software is a particularly challenging issue in both research and industry. Interactions between design decisions have effects on performance and functionality that are difficult to predict. The observation that automated algorithm configuration and parameter tuning tools can simplify this task has led to a new software development paradigm: \textit{Programming by Optimization}~(PbO)~\cite{DBLP:journals/cacm/Hoos12}. Development in the PbO paradigm consists of specifying large design spaces of program component implementations: the onerous task of determining which components work best in a given use case is achieved via automated analysis tools and optimization heuristics.

The standard Programming by Optimization tools operate on design spaces specified in a specialized extension of a target programming language, transformed into the target language by a specialized \emph{weaver} tool. The optimization choices over the combined design space are made by a separate algorithm configuration tool, which has historically been applied to the resulting executable program.

The weaver-based architecture severely limits the applicability of Programming by Optimization. The reliance on markup extensions hinders the adoption of PbO for existing code bases. In addition, the external configuration tools have to operate on the executable via a brittle textual interface (command line arguments), which introduces significant overhead and makes on-line optimization difficult. Despite these shortcomings, no new alternative to weaver tools has been introduced since the initial PbO proposal.

We developed \textsc{ContainAnt}, a software library for Programming by Optimization that addresses these limitations by replacing syntactic extensions and weavers with the Dependency Injection architectural pattern \cite{prassana:dep}. By exploiting a structural correspondence between Dependency Injection and context-free grammars, we obtain a new class of grammar-based evolutionary heuristics suitable for automated algorithm configuration. We determined that these new heuristics significantly outperform existing configuration algorithms on several common configuration tasks and optimization problems, both in terms of solution quality and execution speed (in one case reducing the optimization time from four~hours to 46~seconds).

This paper discusses the theory and implementation of the \textsc{ContainAnt} library and its grammar-based heuristics. Sections~\ref{ssec:di}~and~\ref{ssec:pbo} introduce Dependency Injection and review the existing work on Programming by Optimization. Section~\ref{ssec:bnf} describes the theoretical correspondence between Dependency Injection and optimization problems over context-free grammars, while Section~\ref{ssec:heur} gives novel heuristics for solving the resulting grammatical optimization problems using genetic algorithms and ant colony techniques. The remainder of the paper analyzes five different experiments used to evalute the performance of the \textsc{ContainAnt} heuristics.

\subsection{Dependency Injection} \label{ssec:di}

Object-oriented software provides functionality via multiple interdependent components. Software engineering efforts to handle the problems of dependency instantiation and reference acquisition between these components has led to the widespread adoption of a new type of middleware library, the so-called Dependency Injection (DI) container \cite{prassana:dep}. The term ``Dependency Injection'' was coined by Fowler \cite{FowlerIoC} in 2004 and DI containers have seen increasingly widespread use over the last decade, with popular frameworks including the Java$^\mathtt{TM}$ Spring framework\footnote{\url{http://projects.spring.io/spring-framework} } and Google Guice\footnote{\url{https://github.com/google/guice} }.

Software written using DI inherently exposes highly structured configuration parameters: components are configured by searching over the space of dependencies, without modifying the source code of the components themselves. The traditional operation of a DI container is to perform the wiring between the constructors of dependent objects (also known as the `object graph') by consulting a configuration object or file that contains a list of bindings between abstract types and their constructor arguments. The container then selects a target class and greedily supplies the dependencies to a suitable constructor of the target class. At this point, it is worth noting a significant limitation of some popular DI containers (e.g.\ Guice): configuration is not possible if the object graph contains ambiguities such as a choice of multiple subtypes of an abstract class. As described in detail in Section 
\ref{ssec:heur}, the optimization based approach of \textsc{ContainAnt} removes this limitation.

\subsection{Related  Work}
\label{ssec:pbo}

In their seminal work on Programming~by~Optimization, Hoos \textit{et al.}~\cite{DBLP:journals/cacm/Hoos12} delineated five levels of PbO,
ranging in sophistication from tuning the exposed parameters of an application (Level~$0$) to the use of evidence-based methods for exploring large design spaces as the driving activity for the software design process (Level~$4$).

To realize the higher levels of PbO, they introduced the concept of a PbO-enhanced language: a superset of an existing programming language (e.g.\ PbO-\textsc{Java} or PbO-\textsc{C}) which includes constructs for declaring the possible design choices for parameters and blocks of code. The code written in this markup language is translated into the target language via a syntactic transformation performed by a specialized \emph{PbO weaver} tool, reminiscent of a macro preprocessor. 

The optimized choices (as determined over the combined design spaces on a set of training cases) are made by an external automatic configuration tool. Configuration optimizers have been proposed that use various heuristics, e.g.\ iterated local search \cite{Hutter:2009:PAA:1734953.1734959}, genetic algorithms \cite{Ansotegui:2009:GGA:1788994.1789011} and iterated racing \cite{irace}.
A notable achievement of PbO is the development and use of the \textsc{SMAC} configuration tool \cite{HutHooLey11-smac} to improve upon the state-of-the-art in SAT solving by tuning parameters of the \textsc{Spear} SAT solver. While the majority of configuration optimizers are model-free, \textsc{SMAC} alternates between building a regression model to predict configuration performance and gathering additional performance data based on this model. The regression model is obtained via \emph{random forests}, a method which is known to perform well on categorical variables and also allows quantification of uncertainty.

Search Based Software Engineering (SBSE) \cite{Harman:2012:SSE:2379776.2379787} is the application of heuristic search to various aspects of the software development process, with a strong historical emphasis on software testing. Much recent interest within SBSE has focused on `embedded adaptivity' \cite{Harman:2014:GIA:2593929.2600116}, i.e.\ allowing software developers to delegate the configuration/generation of specified aspects of program functionality to heuristic search procedures \cite{Burles:2015:EDI:2739482.2768423}. Such SBSE activity is therefore strongly aligned with the previously stated goals of PbO, but often with emphasis on a generation process which can respond dynamically to changes in the operating environment of the program. Previous work in this area includes \textsc{Gen-O-Fix} \cite{GenofixTR} and ECSELR \cite{Yeboah-Antwi:2015:EAS:2739482.2768425},
both of which are embedded monitor systems that support search via Evolutionary Computation. \textsc{Templar} and \textsc{Polytope} are two alternative approaches to software component generation: \textsc{Templar} \cite{TemplarSwan2015} provides a `top-down' framework for orchestrating one or more `variation points' generated by Genetic Programming, while \textsc{Polytope} \cite{Swan:2017:evoApplications} uses methods from datatype generic programming to support the `bottom up' generation of individual variation points in source code.

Since Dependency Injection containers already automate a non-trivial part of the Software Engineering process, they provide a natural entry point for the application of heuristic methods from SBSE. 

\section{Grammatical Optimization} \label{ssec:bnf}

Backus-Naur Form (BNF) is a widely adopted syntax for describing context-free languages. For the sake of technical convenience (the ability to have different rewrite rules with identical bodies), we present a slight variation of the usual notion, the labeled~BNF~formalism introduced by Forsberg~and~Ranta~\cite{forsberg:lbnf}. Formally, such a grammar $\mathcal{G}$ consists of the following components:
\begin{itemize}
\item A set of \textbf{terminal symbols}~$\mathcal{G}_T$. These are the literals or words that make up the language.
\item A pointed set of \textbf{non-terminal symbols}~$\mathcal{G}_N$, with a distinguished \textbf{start symbol} $\mathtt{s} \in \mathcal{G}_N$. These categorize the sub-expressions of the language.
\item A set of \textbf{rewrite rules}~$\mathcal{G}_R$. Normally, each rewrite rule has the form $(a,\overline{b})$ where $a \in \mathcal{G}_N$ and $\overline{b}$ is a sequence of symbols from $\mathcal{G}_T \cup \mathcal{G}_N$. Since we are dealing with labeled BNF, rewrite rules have the form $(\ell, a,\overline{b})$ where $\ell$ is a unique \textbf{label}, the \textbf{left-hand side}~$a$ is a non-terminal and the \textbf{right-hand side}~$\overline{b}$ is a finite sequence of symbols from $\mathcal{G}_T \cup \mathcal{G}_N$.
\end{itemize}

At this point, it is customary to introduce the notion of sentence: a sequence of terminal symbols obtained from the start~symbol~$s$ by applying a sequence of rewrite rules, i.e.\ by replacing non-terminals with the right-hand sides of the corresponding rewrite rules. 
Such a sequence of rules can be represented as a rooted tree known as a \emph{derivation tree}. A grammar is unambiguous if each of its sentences has a unique corresponding derivation tree.

In practice, the actual sentences of the language turn out to be immaterial from the perspective of a grammatical optimization problem, so it is simpler to work from a direct definition of derivation trees. Thus we ignore the underlying sentences altogether and inductively define a derivation tree of sort $x \in \mathcal{G}_N$ to consist of the following data:
\begin{itemize}
\item A rewrite rule of the form $(\ell, x, \overline{b})$,
\item A derivation tree of sort $a \in \mathcal{G}_N$ for every non-terminal symbol~$a$ in the sequence~$\overline{b}$.
\end{itemize}

From here on, all derivation trees are assumed to have sort~$\mathtt{s}$~(the start symbol of the grammar $\mathcal{G}$). The set of all such derivation trees is denoted $D(\mathcal{G})$.

Grammars can be specified by listing their rewrite rules in the following format:
\begin{verbatim}
Label. <LHS> ::= RHS
\end{verbatim}
where angled brackets are used to distinguish between terminals and non-terminals.

\noindent
Two elementary examples follow:
\subsubsection{Binary Strings}

The grammar of binary strings is given by:
\begin{align*}
G_T &= \left\{\mathtt{0}, \mathtt{1}, \mathtt{e} \right\}, \\
G_N &= \left\{\mathtt{s} \right\}, \\
G_R &= \left\{ (\mathtt{0},\mathtt{s},\mathtt{0s}), (\mathtt{1},\mathtt{s},\mathtt{1s}), (\mathtt{e},\mathtt{s},\mathtt{e}) \right\}.
\end{align*}

Using the shorthand defined above, the rewrite rules could also be written as
\begin{verbatim}
0. <s> ::= 0 <s>
1. <s> ::= 1 <s>
e. <s> ::= e
\end{verbatim}

Derivation trees for this grammar correspond to finite sequences of binary digits (with \texttt{e} being the terminating character). A grammar of strings over any given finite alphabet can be defined analogously.

\subsubsection{Finite Sets}

Any finite set~$S$ gives rise to a grammar by~setting
\begin{align*}
G_T &= S, \\
G_N &= \left\{\mathtt{s} \right\}, \\
G_R &= \left\{ (x,\mathtt{s},x)\:|\: x \in S \right\}.
\end{align*}

The derivation trees of this grammar are in bijective correspondence with elements of the set~$S$.

\subsection{Problem Statement}

An instance of the \textbf{grammatical optimization problem} is given by the following data:
\begin{itemize}
\item A grammar $\mathcal{G}$ and
\item An \textbf{objective~function} $f: D(\mathcal{G}) \rightarrow \mathbb{R}$ defined on the derivation trees of the grammar $\mathcal{G}$.
\end{itemize}

Without loss of generality, we assume that our goal is \textit{maximizing} the objective function, i.e.~solving the grammatical optimization problem consists of finding a globally optimal derivation tree: 
$$x^* = \mathrm{arg} \mkern-18mu \max_{x \in D(\mathcal{G})\ \ \ \ \ } \mkern-18mu f(x).$$

The definition above is extremely general: indeed, \textit{every} discrete optimization problem can be reduced to the grammatical optimization problem over the grammar of binary strings.

\subsection{Semantics}

One can reduce an optimization problem instance with candidate solutions~$S$ and objective function~$f: S \rightarrow \mathbb{R}$ to an instance of the grammatical optimization problem by giving an encoding grammar $\mathcal{G}$ and a surjective function
$$k: D(\mathcal{G}) \rrightarrow S$$
with surjectivity ensuring that every candidate solution is described by at least one sentence of the language.

To forbid \emph{ad-hoc} encodings (e.g.\ the~encoding of any discrete optimization problem into the grammar of binary strings discussed above), one should think of the function~$k$ as giving a \textit{semantics} to the sentences of the language defined by the grammar~$\mathcal{G}$. From here on, we demand that the semantics be compositional: the meaning of a derivation tree should be given in terms of the meanings of its parts (direct subtrees). The compositionality requirement provides a formal counterpart to the intuitive desideratum that the structure of the grammar be related to the structure of the search space~$S$, without ruling out any interesting grammatical representations.

We will shortly see that both dependency injection and the algorithm configuration problem have sensible, compositional representations as instances of the grammatical optimization problem. What's more, the same holds for many problems of interest in both continuous and combinatorial optimization.

\subsection{Rosetta Stone}

Analyzing the process of dependency injection leads to a powerful ``dictionary'' correlating the terminology of grammars with the terminology of object-oriented programming.
If the goal is to instantiate an object of some given class $C$, one first has to find a constructor of $C$ (if the class has no constructors, instantiation is impossible). In turn, the selected constructor will expose zero or more classes as dependencies. If the selected constructor $c()$ has no dependencies, the object can be instantiated directly by calling $c()$. However, if there are one or more dependencies $D_1, D_2, \dots$, one has to recursively instantiate objects $d_1,d_2,\dots$ compatible with the given classes before calling $c(d_1,d_2,\dots)$ to instantiate an object of class~$C$.

Now, let $\mathcal{G}$ be a grammar. To construct a derivation tree of some given sort $\mathtt{s} \in \mathcal{G}_N$, one starts by choosing a rewrite rule with left-hand side $\mathtt{s}$ (no suitable tree can exist in the absence of such a rule). If the right-hand side of the chosen rewrite rule contains no non-terminals, the construction is finished. However, if the right-hand side contains one or more non-terminals $n_1,n_2,\dots \in \mathcal{G}_N$, one has to recursively construct a derivation tree for each sort $n_i$ before constructing the derivation tree for the target sort $\mathtt{s}$. 

The structure of the algorithms for dependency injection and derivation tree construction (Algorithms~\ref{alg:rosetta-object}~and~\ref{alg:rosetta-grammar}) turn out to be nigh-identical. This suggests an analogy between dependency injection and grammatical optimization, with classes corresponding to non-terminals, constructors corresponding to rewrite rules and constants corresponding to terminals. Thus, the grammatical rewrite rule corresponding to the constructor (\textsc{Java} syntax) 
\begin{align*}
\mathrm{T}\ ctor(\mathrm{T}1\ a1,\ \mathrm{T}2\ a2,\dots)
\end{align*}
under this assignment is simply
\begin{verbatim}
    ctor. <T> ::= ctor <T1> <T2> [...]
\end{verbatim}

With this correspondence in mind, we can now recast dependency injection as a grammatical decision/optimization problem.

Given a grammar $\mathcal{G}$, deciding whether $D(\mathcal{G})=\emptyset$ amounts to solving a dependency injection problem. The correspondence gives rise to a semantics assigning the constructed object to each derivation tree of the grammar. In the sequel, this is referred to as the \textit{usual semantics}.

\begin{algorithm}
\caption{Class instantiation using Dependency Injection}
\lstset{language=Pascal,
basicstyle=\footnotesize
}
\begin{lstlisting} 
function instantiate(t: Class)
  for c in t.constructors
    {try to construct each argument recursively}
    for i := 0 to c.arguments.length
      args[i] := instantiate(classOf(a))
    end for
    {if recursive calls succeed}
    if !args.contains(null)
      return c(args) {call constructor}
    end if
    {else try the next constructor}
  end for
  return null
end function
\end{lstlisting}
\label{alg:rosetta-object}
\end{algorithm}

\begin{algorithm}
\caption{Recursive Derivation Tree Construction}
\lstset{language=Pascal,
basicstyle=\footnotesize
}
\begin{lstlisting} 
function construct(t: Sort)
  for c in t.rewriteRules
    {try to construct each subtree recursively}
    for i := 0 to c.nonterminals.count
      subtrees[i] := construct(sortOf(i))
    end for
    {if recursive calls succeed}
    if !subtrees.contains(null)
      return Tree(c,subtrees)
    end if
    {else try the next rewrite rule}
  end for
  return null
end function
\end{lstlisting}
\label{alg:rosetta-grammar}
\end{algorithm}

\section{Heuristics} \label{ssec:heur}

\textsc{ContainAnt} is, first and foremost, a Dependency Injection library. 
In order to be as widely applicable as Programming~by~Optimization, the default heuristics of \textsc{ContainAnt}~cannot be problem-specific: they have to operate at the level of problem descriptions. Metaheuristics that only exist as nature-inspired metaphors or informal algorithm templates (i.e.\ without the ability  to automatically transform a problem specification into a working implementation) are insufficient for this. These requirements leave us with a rather small class of suitable metaheuristics, which we now describe.

\subsection{Genetic Programming: \textsc{GrEvo}}

Incorporating context-free grammars into genetic programming was proposed by Ryan~\textit{et al.}~\cite{ryan:grevo}. Their seminal work on~\textit{Grammatical~Evolution} allowed the elimination of the closure requirement, a major drawback of untyped Genetic Programming, which required all functions to be able to accept as input the outputs of all other functions. The genotypes are numerical sequences, translated into sentences of a BNF grammar using the mapping of Algorithm~\ref{alg:grevo-geno}. Transcribed into the derivation tree formalism of Section~\ref{ssec:bnf}, the genotypes encode the choice of rewrite rule at each recursive step of the derivation tree construction~(Algorithm~\ref{alg:rosetta-object}).

\begin{algorithm}
\caption{\textsc{GrEvo} Genotype-Phenotype Mapping}
\lstset{language=Pascal,
basicstyle=\footnotesize
}
\begin{lstlisting} 
function getPhenotype(g: List[Int], t: Sort)
  {choose rewrite rule based on genotype}
  c := t.rewriteRules[g.head]
  g := g.tail
  for i := 0 to c.nonterminals.count
    subtrees[i] := getPhenotype(g, sortOf(i))
  end for
  return Tree(c,subtrees)
end function
\end{lstlisting}
\label{alg:grevo-geno}
\end{algorithm}

Since Grammatical~Evolution allows the generation of syntactically correct sentences in an arbitrary language, its implementations are not tied to any specific problem, and are able to operate on any formal grammar specification. Grammatical~Evolution remains the most popular metaheuristic of its kind, generally outperforming derivate algorithms such as Grammatical Swarm~\cite{oneill:swarm}.

The \textsc{ContainAnt} distribution includes an implementation of the Grammatical Evolution metaheuristic with fixed-length genotypes for solving the grammatical optimization problem. This implementation is henceforth called~\textsc{GrEvo}. The performance analysis~(Section~\ref{ssec:anal}) shows that some characteristics of \textsc{GrEvo}, such as its premature convergence and poor locality, make it suboptimal for tackling the grammatical optimization problem. This limitation motivates the novel grammar-based heuristic introduced below.

\subsection{Ant Programming: \textsc{GrAnt}}

Ant colony optimization methods are the main alternative to Genetic Programming for the automated production of computer programs via stochastic search. Ant Programming based on BNF grammars has been investigated by Keber~and~Schuster~\cite{keber:gap} under the name Generalized Ant Programming (GAP) in the context of option pricing, and later by Salehi-Abari~and~White~\cite{salehi:egap} for general automatic programming~(EGAP). The development of these heuristics led to what has been called an ``up-hill battle'' between the two methods, while genetic programming was found to be statistically superior to~EGAP~\cite{salehi:uphill}.

Here, we describe a novel ant colony algorithm~(\textsc{GrAnt}) for solving the grammatical optimization problem that significantly outperforms Grammatical~Evolution on diverse optimization problems. The new heuristic is based on the MIN-MAX Ant System\cite{stutzle:mmas}, but differs from previous Ant Programming algorithms on two key points:
\begin{enumerate}
\item The pheromone levels (associated with rewrite rules) are bounded between a minimum and maximum pheromone value. However, the maximum is treated as a soft bound that can be changed by specific events over the course of the search.
\item Each ant constructs a complete derivation tree in a depth-first, targeted fashion (cf.~EGAP'a use of partial sentences and non-terminals).
\end{enumerate}

\textsc{GrAnt}~(Algorithm~\ref{alg:grant}) maintains a pheromone table, holding a pheromone level lying between a hard minimum level $\tau_{min}$, and a soft maximum $\tau_{max}$ for every rewrite rule.
A search iteration begins with each ant constructing a derivation tree of the target sort.The construction proceeds by recursively choosing rewrite rules using simple pheromone-proportional selection. The fitness of the constructed trees is calculated, pheromones are updated by applying evaporation. The iteration-best ant is allowed to deposit pheromones by adding the fitness value to the pheromone level of each rewrite rule used in the derivation.  If the iteration-best fitness ever exceeds $\tau_{max}$, then $\tau_{max}$ is updated to the higher value. The motivation for this behavior is assigning more weight to pheromone increases caused by finding fit solutions vs. pheromone buildup caused by repeatedly exploring an area of the search space. As an additional benefit, this eliminates the need for normalizing the amount of pheromones on the edges (shaking). Upon reaching the stopping condition, the algorithm returns the overall best solution found.

\begin{algorithm}
\caption{\textsc{GrAnt} Heuristic}
\lstset{language=Pascal,
basicstyle=\footnotesize
}
\begin{lstlisting}
function grant(t: Sort)
  while (! stopped)
    solv := construct(p,t) {wlog 1 ant}
    iter := fitnessOf(solv)
    evaporatePheromone()
    for rule in solv
      addPheromone(rule, iter)
    end for
    {update max pheromone}
    if iter > tau_max then
      tau_max := iter
    end if
    {update best solution}
    if fitnessOf(best) > iter then
      best := solv
    end if
  end while
  return best
end function

{recursive path construction}
function construct(p: Pheromones, t: Sort)
  {pheromone-proportional rule selection}
  c := p.select(rewriteRules(target))
  {construct subtree for each non-terminal
  of the selected rule}
  for i := 0 to c.nonterminals.count
    subtrees[i] := construct(p, sortOf(i))
  end for
  return Tree(c, subtrees)
end function
\end{lstlisting}
\label{alg:grant}
\end{algorithm}

\section{Implementation}

\textsc{ContainAnt} is implemented as a Dependency Injection library for the \textsc{Scala} programming language. The statically typed, object-oriented nature of \textsc{Scala} makes it well-suited for Dependency Injection, and its run-time reflection facilities tremendously simplify the \textsc{ContainAnt} architecture. Moreover, \textsc{Scala} runs on the Java Virtual Machine, allowing the library to work with code bases written in any JVM language (including \textsc{Clojure} and \textsc{Java}).

\textsc{ContainAnt}'s job is assembling objects and object graphs. In effect, the library takes over object instantiation. Instead of using the \texttt{new} keyword with a constructor to instantiate classes, the programmer requests an instance of a given class from \textsc{ContainAnt} (\texttt{ContainAnt create[ClassName]}). The container then heuristically determines what to build by resolving dependencies, choosing appropriate constructors and wiring everything together.

To take advantage of the heuristic capabilities, the programmer has to supply an objective function. With the exception of this objective function, the configuration of \textsc{ContainAnt} is modeled on Google's popular \textsc{Guice} dependency injection library. The programmer provides a \texttt{Module} (a plain object implementing a marker trait) containing the constructors and helper functions to be used during Dependency Injection. If the software to be optimized uses Dependency Injection, these modules will already be present, ready to be used by \textsc{ContainAnt}. This is in strict contrast with the weaver~approach to Programming by Optimization: weaver rules are not present in programs that were not designed with the corresponding PbO toolset in mind.

\textsc{ContainAnt} parses module specifications using \textsc{Scala}'s reflection capabilities, turning the Dependency Injection problem into a grammatical optimization instance. Our analysis~(Section~\ref{ssec:anal}) indicates that the default \textsc{GrAnt} search heuristic suffices to solve many optimization and algorithm configuration problems without problem-specific tuning. This means that using \textsc{ContainAnt} does not require the practitioner to deal with grammars, or even being aware of the heuristics working ``under the hood''.

Since \textsc{ContainAnt} acts like an ordinary dependency injection container, taking over the instantiation of objects and resolution of dependencies, it need not distinguish between off-line and on-line~adaptive optimization: the distinction can be made by using an embedded wrapper to select between `construct on first use' or dynamic/periodic reconstruction \cite{Burles:2015:EDI:2739482.2768423}.

There are no major obstacles to turning the container into a drop-in replacement for \textsc{Guice} by implementing the complete \textsc{Guice} API, thus making PbO immediately available to hundreds of enterprise software projects. This is possibly the most important application of the correspondence detailed in Section~\ref{ssec:bnf}, and the main future target of \textsc{ContainAnt}~development.

\section{Case Studies}

To demonstrate the general behavior of \textsc{ContainAnt}~and~\textsc{SMAC} \cite{HutHooLey11-smac}, and to compare the performance of their heuristics, we implemented two classical optimization problems (Branin function, Subset Sum) and three algorithm configuration problems ($D$-ary heaps, skiplists and syntax highlighting). For comparison purposes, one problem of each class was also implemented for use with \textsc{SMAC}. In this section, we offer a detailed look at each problem, followed by a performance comparison showing that \textsc{GrAnt}~significantly outperform the other heuristics in all but one of these problems.

\subsection{Classical Problems}
\subsubsection{Branin Function}

In this first case study, we compare \textsc{ContainAnt} with \textsc{SMAC} on a global optimization problem. The goal is to minimize the value of the \textit{Branin function} on a given bounded subset of the Euclidean plane.
The Branin function (introduced by Dixon~and~Szeg\H{o} in their traditional optimization test suit~\cite{dixon:global}) has long been a popular benchmark for continuous optimization heuristics. The function has the form
\begin{align*}
\mathrm{branin}(x_1,x_2) =& \left( \left(x_2 - \frac{5.1}{4\pi^2}\right) x_1^2 + \frac{5}{\pi}x_1 - 6 \right)^2 + \\
& 10\left( 1 - \frac{1}{8\pi} \right) \cos(x_1) + 10
\end{align*}
with the domain restricted so that $x_1 \in [-5, 10]$ and $x_2 \in [0, 15]$. There are three global minima on this domain, each with value $\sim 0.397 = 2.48^{-1}$.

The Branin function provides an ideal context for comparing the behavior and the performance of \textsc{SMAC} and \textsc{ContainAnt}, since the \textsc{SMAC}~distribution already includes a configuration for optimizing the Branin function in one of the default example scenarios.

There are many practical techniques for representing a continuous solution space as a BNF grammar. The most intuitive way is including a sufficiently fine ``uniform grid'' of constants from the domain as terminals of the grammar. Alternatively, the grammar of binary strings presented in Section~\ref{ssec:bnf} can represent every dyadic fraction in a compact interval. Dyadic fractions form a dense subset of the interval and provide arbitrary-precision approximations to any given number. We decided to go with the former, more intuitive grammar for this experiment. The result is a large grammar with many terminals, but one that aligns very well with \textsc{SMAC}'s solution representation, thereby ensuring that both heuristics explore search spaces of the same size, which leads to a completely fair comparison.

\subsubsection{Subset Sum}
Given a finite set of integers $S \subseteq \mathbb{Z}$ and target~number~$c \in \mathbb{Z}$, is there a subset~$I \subseteq S$ such that 
$$\sum_{i \in I} i = c\:?$$

Known as ``subset sum'', this is one of the ur-examples of an NP-complete decision problem. Recast as an optimization problem, we will attempt to maximize the function $f(I) = \left|c - \sum I\right|^{-1}$ with $f(I) = 2$ if $\sum I = c$. We use two subset sum benchmark instances (P01 and P03) from Burkardt's Scientific Computing Dataset~\cite{burkardt:data} for this case study.

The grammar for this instance consists of the finite grammar generated by the numbers in~$S$, along with the following generic rewrite rules for constructing sets of numbers:
\begin{verbatim}
empty. <Set> ::= empty
add. <Set> ::= add <Int> <Set>
\end{verbatim}
with the obvious compositional semantics
\begin{align*}
k(\mathtt{empty}) &= \emptyset \\
k(\mathtt{add}\:y\:z) &= \left\{k(y)\right\} \cup k(z)
\end{align*}
Notice that the argument-passing system of \textsc{SMAC} would not be capable of supplying arguments of this complexity. The experiment is limited to the \textsc{ContainAnt} heuristics, with 100 runs and the heuristics capped at 1000 objective function evaluations.

\subsection{Programming by Optimization}

\subsubsection{D-ary Heaps}\label{ssec:dheap}

A \emph{min-heap} (resp.~max-heap) structure is a rooted tree in which every node has a value larger (smaller) than the value of its parent. A $D$-ary heap is a heap structure built on a complete $D$-ary tree. The familiar binary heaps are $D$-ary heaps with $D=2$. General $D$-ary heaps allow faster key update operations than the binary case --- $O\left(\log_D n\right)$ vs. $O\left(\log_2 n\right)$. This makes $D$-ary min-heaps (resp.~max-heaps) appropriate for algorithms where decrease (increase) operations are more common than minimum (maximum) extraction.

Generalizing the binary case, the underlying tree can always be implemented as an array, with the children of the $i$th node placed at indices $iD + 1, iD + 2, \dots, iD + D$. This implementation strategy improves cache efficiency and enables random access. There is a performance trade-off, however: the array will eventually fill up, triggering an expensive resize operation.

A $D$-ary heap data structure implemented with arrays has three parameters: the initial size of the array, the expansion factor of the resize operation, and (of course) the arity~$D$. The optimal values of these parameters depend on the expected number of values to be stored in the structure, as well as the expected distribution of decrease/increase and minimum/maximum extraction operations.

\begin{figure}[!t]
\centering
\includegraphics[width=2.5in]{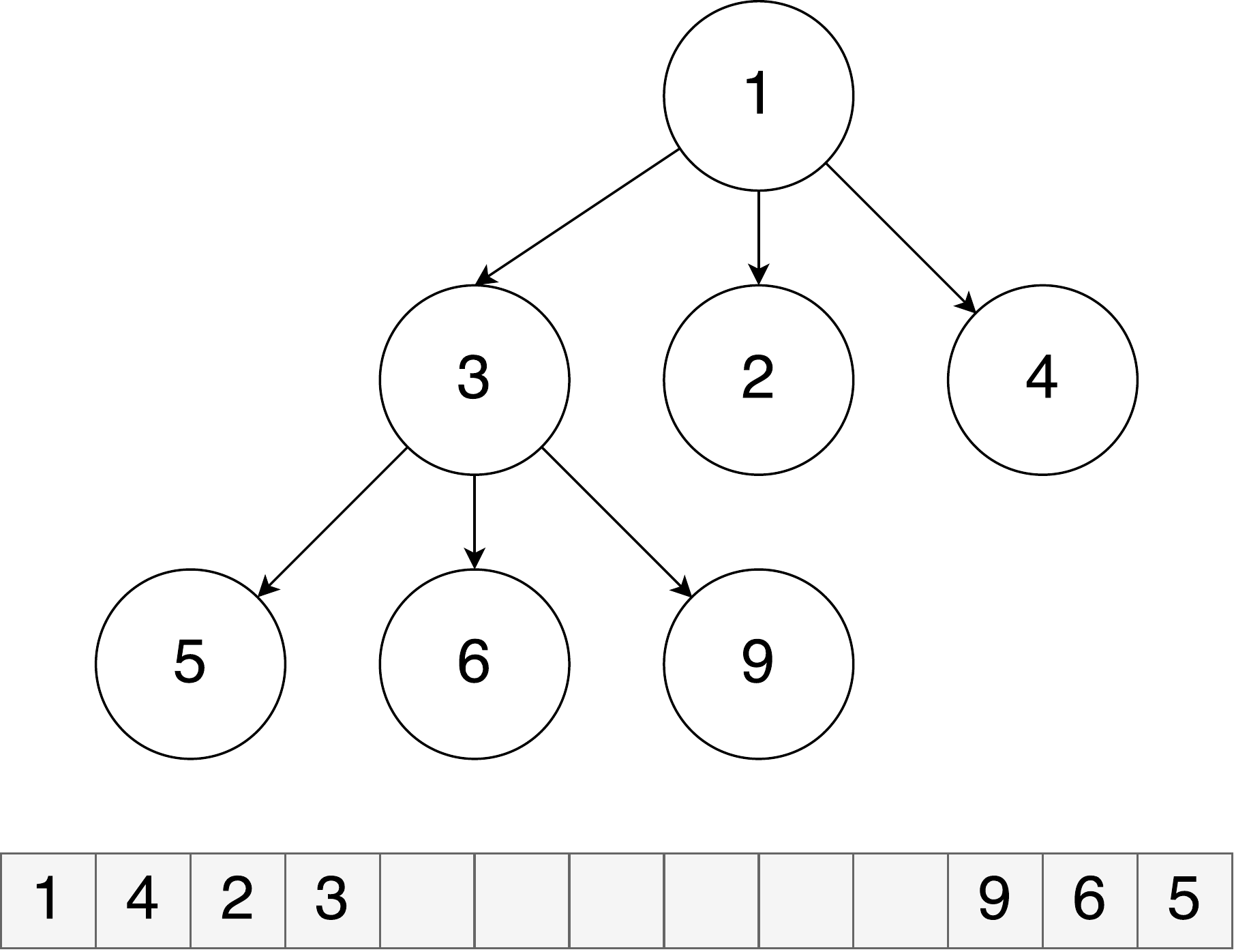}
\caption{A ternary min-heap and its array representation.}
\label{fig:dheap}
\end{figure}

The optimization of $D$-ary heaps was implemented by~Hoos~and~Hsu as a test instance for the original Programming~by~Optimization proposal. The original code is written in an extended dialect of the \textsc{Java} programming language, designed for use with a PbO~weaver. The weaver-specific declarations have to be factored out into constructor arguments - a mere three lines of changes, one for each parameter described above. The resulting standard \textsc{Java} is directly usable by \textsc{ContainAnt}.

The grammar for the data structure configuration problem consists of the constructor for the dynamic heap class as the only proper rewrite rule; there are classes and constants for heaps, their arities, expansion factors and initial sizes, all of them equipped with their usual semantics. The objective function counts the number of accesses to the underlying array (with each resize operation counting as two accesses for each index, in line with the usual amortized analysis for array lists) under a given test load. Evaluating the objective function for this task is very expensive, so the experiment is limited to 10 runs, with the heuristics capped at 1000 objective function evaluations.

\subsubsection{Skiplists}

Skiplists are a probabilitistic alternative to balanced binary search trees \cite{pugh:skiplist}. Skiplists are essentially ordered linked lists where each node may contain multiple forward links. In the familiar linked list, a node consists of a value (a piece of data) and a link to the next node. Nodes in a skiplist contain a whole hierarchy of links, each one pointing to a farther subsequent node than the one below it. These auxiliary links provide an ``express lane'' for navigating the structure and can be exploited to implement all three dictionary operations (insertion, lookup and deletion of values) with logarithmic expected time complexity. Thus, the performance of skiplists is comparable to that of balanced binary search trees.

Skiplists are parametrized by two numeric values: the transition probability $p\in \mathbb{R}$ and the maximal height of the hierarchy~$h\in \mathbb{N}$. To find a given value $v$ in a skiplist, start by following the highest level links of the hierarchy, advancing until either $v$ is encountered, or the value of the next node is greater than $v$. In the latter case, continue the search by following links one level down in the hierarchy.
To insert a given value $v$ into a skiplist, start by finding its location using the method described above. Create a node for storing $v$. Now, generate a uniform random real $x \in [0,1]$ and link the newly created node to its neighbors in level~$\ell$ of the hierarchy if and only if $p^k > \ell$. When $p=0.5$, one can intuitively think of this process as a series of coin flips. If you get heads, you link the node to its neighbors on level $\ell$ of the hierarchy, then repeat the procedure on level $\ell+1$. If you get tails or reach the maximum height $\ell = h$, the insertion operation ends.

Instead of having a fixed parameter~$p$, where the probability of inserting a value into level $k$ of the hierarchy is always $\frac{1}{p^{-k}}$, one can consider a more general skiplist architecture, where this probability is given by $\frac{1}{P_k}$, where~$P: \mathbb{N} \rightarrow \mathbb{R}_+$ is an arbitrary monotone sequence. In the experiment, we will focus on three different types of sequences:
\begin{itemize}
\item \textbf{Geometric}: $P_k = a^k$ for some $a > 1$,
\item \textbf{Arithmetic}: $P_k = a + k$ for some $a > 0$ and
\item Sums of the previous two types.
\end{itemize}
Hence, our skiplists will have two parameters: the maximum height~$h$, and the probability sequence~$P$. The expected time complexity of lookups is independent of the distribution of the values~\cite{motwani:randomized}. However, the optimal choices of the parameters~$P$~and~$h$ do depend on the expected number of items to be stored in the skiplist. Skiplists are often stored in a distributed fashion, where the optimal configuration may further depend on variables such as network latency, giving rise to an on-line data structure configuration problem.

\begin{figure}[!t]
\centering
\includegraphics[width=2.5in]{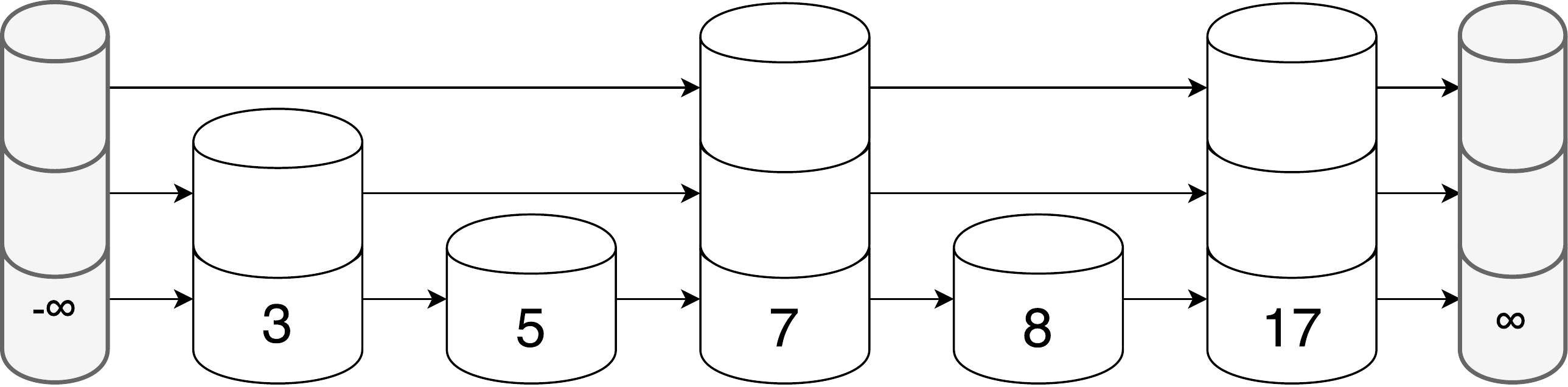}
\caption{Skiplist with a three-layer hierarchy.}
\label{fig:skiplist}
\end{figure}

\textsc{ContainAnt} is readily able to solve this parameter tuning problem --- indeed, we have already evaluated this capability on a much larger search space in Section~\ref{ssec:dheap}. However, we can use optimization to explore a more interesting search space by considering a generalized variant of skiplists. 

Let $\mathtt{p}$ denote the (non-terminal corresponding to) the class of integer sequences. The grammar for this data structure configuration problem has a rewrite rule corresponding to the constructor of the skiplist class, as well as three special rewrite rules for constructing the probability sequences:
\begin{verbatim}
geom. <Prob> ::= geom <Double>
arit. <Prob> ::= arit <Double>
sum. <Prob'> ::= sum <Prob> <Prob>
\end{verbatim}

The compositional semantics assigns
\begin{align*}
k(\mathtt{geom}\:y) &= (1,k(y),k(y)^2,k(y)^3,\dots) \\
k(\mathtt{arit}\:y) &= (1,1+k(y),1+2k(y),\dots) \\
k(\mathtt{sum}\:y\:z) &= k(y) \oplus k(z)
\end{align*}
where the symbol $\oplus$ denotes the termwise sum of two sequences. As in the other grammars, there are constructors for skiplists and constants for the numerical parameters, all of them equipped with their usual semantics. This shows that the grammatical approach can conveniently represent sophisticated search spaces that would be difficult and sometimes impossible to describe via \textsc{SMAC}'s text-based configuration files.
The objective function fills the skiplist structure with 1000 random values, and performs 100 random lookups, measuring the total number of comparisons performed. All heuristics are capped at 100 objective function evaluations. The search is fast enough to make 100 runs of the experiment feasible.

\subsubsection{Syntax Highlighting}\label{ssec:syntax}

This final case study serves to showcase a practical use case for Programming~by~Optimization in general and \textsc{ContainAnt} in particular: the creation of software with search-based ``dynamic adaptive'' features. Our minimal example is a syntax highlighter that automatically adjusts itself to different display environments. The potential applications include battery-saving color schemes compatible across different devices (using the technique of Burles~\textit{et al.}~\cite{burles:energy} to incorporate energy consumption into the objective function) and schemes that remain readable when transplanted to different environments (e.g. embedded into social media or displayed by the fixed background color ``webview'' of a mobile application).

\textsc{Agda} is an increasingly popular dependently typed programming language designed by Ulf~Norell \cite{norell:agda}. The \textsc{Agda}~compiler can generate documentation web pages which include the navigable, syntax-highlighted source code of the compiled software. Unfortunately, the default color scheme for the syntax highlighting is unreadable on dark backgrounds, which causes problems when embedding the generated documentation into a larger website. 

Our test program generates a readable color scheme for \textsc{Agda} documentation given a target background color as input. The program consists of little more than a naive fitness function quantifying the readability of a color scheme by penalizing low contrast and by rewarding color schemes based around a small number of complementary colors. All of the search is relegated to either~\textsc{SMAC}~or~\textsc{ContainAnt}. The former requires a configuration file with 27 categorical variables, each with 27 options. In addition, about 100 lines of boilerplate code had to be written for handling command line arguments and interfacing with~\textsc{SMAC}. For \textsc{ContainAnt}, the grammar specification, consisting of the constructors for the \texttt{ColorScheme} and \texttt{RGBValue} classes, takes 37 lines altogether. The heuristics are capped at 1000 objective function evaluations.

\subsection{Analysis}\label{ssec:anal}

All experiments were performed on the following system:
\begin{itemize}
\item \textbf{CPU}: Intel Xeon E5-2676 clocked at 2.40GHz with 30~MB Level 3 cache,
\item \textbf{RAM}: 1019280k total,
\item \textbf{Swap}: disabled,
\item \textbf{JVM version}: 1.8.0\_121.
\end{itemize}

The data and code that actually conducted this analysis are published in the companion GitHub repository\footnote{ \url{https://github.com/zaklogician/ContainAnt} } of the article. 
The \textsc{ContainAnt} implementation is deterministic, and the repository bundles a convenient build script, allowing anyone to execute the same analysis and replicate/duplicate our results.

The performance of the heuristics was compared on three variables: 
\begin{enumerate}
\item The mean quality (\textit{avg}) achieved by the best~of~run solution returned by the heuristic, averaged over all runs.
\item The optimum quality (\textit{max}) achieved by the best~of~run solution returned by the heuristic, taken over all runs.
\item The variance\footnote{Important for on-line optimization, where the heuristics will be run a large number of times. A technique with high mean but low variance may well lose out to another technique with lower mean but high variance over a large number of runs.} (\textit{var}) of the quality achieved by the best~of~run solutions, taken over all runs.
\end{enumerate}

The significance of the differences between the performance of the top heuristics is checked using the nonparametric protocol of Wineberg~and~Christensen~\cite{wineberg:stat}. The final $p$-values are reported in Table~\ref{tbl:results}.

Each experiment is performed with a fixed number of runs (that number depending on the case study, as explained in the respective subsections). Our goal is to pick the technique that achieves the solution of the highest quality possible, given a single run with a fixed budget of ``computational effort''. To ensure fair comparison, we need to limit the number of objective function evaluations identically for all heuristics. For the constructive heuristics (random search and~\textsc{GrAnt}), this can easily be achieved by capping the number of iterations. For \textsc{GrEvo}, the number of evaluations depends only on the population size and the number of generations, allowing us to limit the number of evaluations by capping the product of these two parameters. \textsc{SMAC} has a mechanism for imposing this cap directly via the configuration file.

The \textsc{GrEvo} heuristic has some tunable (hyper)-parameters, including population size and the number of generations. We hand-selected the best-performing ratio of these parameters from the set $\left\{ (100:10), (40:25), (25:40), (10:100) \right\}$ separately for each case study. \textsc{ContainAnt} is capable of tuning the hyper-parameters of its own heuristics. In principle \textsc{ContainAnt} could be used as its own hyper-heuristic to self-improve \textsc{GrEvo}. We experimented with these capabilities during the early days of development. However, we abandoned this avenue once evidence emerged that significant improvement to these parameters would not be possible within the constraints of the case studies (see~the~paragraph dedicated to \textsc{GrEvo} below).

Table~\ref{tbl:results} summarizes the results achieved by~\textsc{SMAC} and the \textsc{ContainAnt}~heuristics on all five case studies presented above.

\begin{table*}[!t]
\centering
\caption{Performance of~\textsc{SMAC}, \textsc{ContainAnt}~heuristics and random search on the five case studies.}
\label{tbl:results}
\begin{tabular}{lllll}
\textbf{Branin:} & {\ul \textbf{GrAnt}} & {\ul \textbf{GrEvo}} & {\ul \textbf{Rand.}} & {\ul \textbf{SMAC}} \\
max:               & 2.48                 & 1.55                 & 2.45                 & 2.48                \\
avg:               & 1.80                 & 0.87                 & 1.37                 & 1.47                \\
var:               & 0.24                 & 0.08                 & 0.32                 & 0.35                \\
p:                 & \textless.001        &                      &                      &\\

\textbf{Subset Sum P02:} & {\ul \textbf{GrAnt}} & {\ul \textbf{GrEvo}} & {\ul \textbf{Rand.}} & {\ul \textbf{SMAC}} \\
max:               & 2.00                 & 1.00                 & 1.00                 & ~-                \\
avg:               & 0.91                 & 0.65                 & 0.03                 & ~-                \\
var:               & 0.45                 & 0.22                 & 0.02                 & ~-                \\
p:                 & \textless.001        &                      &                      &\\

\textbf{Subset Sum P03:} & {\ul \textbf{GrAnt}} & {\ul \textbf{GrEvo}} & {\ul \textbf{Rand.}} & {\ul \textbf{SMAC}} \\
max:               & 2.00                 & 0.00                 & 0.00                 & ~-                \\
avg:               & 0.38                 & 0.00                 & 0.00                 & ~-                \\
var:               & 0.62                 & 0.00                 & 0.00                 & ~-                \\
p:                 & \textless.001        &                      &                      &\\
\textbf{DHeap:} & {\ul \textbf{GrAnt}} & {\ul \textbf{GrEvo}} & {\ul \textbf{Rand.}} & {\ul \textbf{SMAC}} \\
max:               & 46801                & 46801                & 46801                & ~-                \\
avg:               & 46801                & 46752                & 46594                & ~-                \\
var:               & 0                    & 10671                & 52919                & ~-                \\
p:                 & 0.168                &                      &                      &\\
\textbf{Skiplist:} & {\ul \textbf{GrAnt}} & {\ul \textbf{GrEvo}} & {\ul \textbf{Rand.}} & {\ul \textbf{SMAC}} \\
max:               & 0.33                 & 0.25                 & 0.27                 & ~-                \\
avg:               & 0.28                 & 0.25                 & 0.25                 & ~-                \\
var:               & 0.01                 & 0.00                 & 0.00                 & ~-                \\
p:                 & \textless.001        &                      &                      &\\
\textbf{Syntax H. Blue:} & {\ul \textbf{GrAnt}} & {\ul \textbf{GrEvo}} & {\ul \textbf{Rand.}} & {\ul \textbf{SMAC}} \\
max:               & 37.82                & 38.02                & 37.57                & ~-               \\
avg:               & 34.36                & 34.50                & 32.18                & ~-                \\
var:               & 5.85                 & 8.73                 & 7.46                 & ~-                  \\
p:                 & 0.663                &                      &                      &\\
\textbf{Syntax H. Yellow:} & {\ul \textbf{GrAnt}} & {\ul \textbf{GrEvo}} & {\ul \textbf{Rand.}} & {\ul \textbf{SMAC}} \\
max:               & 34.92                & 34.68                & 33.85                & 34.44                \\
avg:               & 31.51                & 32.33                & 29.22                & 30.92             \\
var:               & 4.02                 & 3.30                 & 5.67                 & 5.22                \\
p:                 & 0.082                &                      &                      &
\end{tabular}
\end{table*}

\subsubsection*{GrAnt}
The \textsc{GrAnt} heuristic significantly outperformed all others in the majority of experiments. The only exception is the syntax~highlighting study, where \textsc{GrEvo} systematically had the highest nominal mean. However, hypothesis testing reveals that the differences are not significant. \textsc{GrAnt} is the only heuristic to perform equally well across both combinatorial~optimization and algorithm~configuration problems, and the only one to find globally optimal solutions to both the Branin~function and both subset sum instances.
\subsubsection*{GrEvo}
The poor performance of the \textsc{GrEvo} heuristic, consistent across parameter settings, is crying out for an explanation. Our investigation suggests that the main culprit may be early loss of diversity (visible in the Skiplist~study, where the algorithm converges in a mere five generations), caused by the fact that the first few elements of the genome have a disproportionately high influence on the phenotype in Grammatical~Evolution~\cite{rothlauf:locality}. Increasing the population size is not possible without moving beyond the strict computational bounds of our case studies, rendering~Grammatical~Evolution unsuitable for many real-time applications. Solving this issue could be an avenue of further research.
\subsubsection*{SMAC}
As expected, the quality of the results returned by~\textsc{SMAC} significantly outperformed random search in all cases. However, the average quality lingered beneath that of~\textsc{GrAnt} in the case of the Branin function (although the best solution for the Branin function was globally optimal) and beneath both \textsc{ContainAnt} heuristics in the algorithm configuration case. Another major issue is speed: \textsc{SMAC} spends over four hours on latter problem, while the \textsc{ContainAnt} heuristics finish both in 46~seconds.

\section{Conclusion}

Dependency Injection can be used to improve the existing weaver-based Programming by Optimization tools. We have described a library that implements several grammatical optimization metaheuristics, including a novel Ant Programming approach. The library provides better support for Programming by Optimization than specialized language extensions and weaver tools, while doing away with several limitations such as difficulties with on-line optimization. 

Furthermore, regarding Dependency Injection as an instance of a grammatical optimization problem leads to a whole new class of heuristics for automatic algorithm configuration. The proposed grammatical Ant Programming heuristic \textsc{GrAnt} significantly outperforms existing algorithms on five problems of interest, in one case reducing a four hour long \textsc{SMAC} optimization task to 46 seconds while significantly improving on the solution quality.

Programming by Optimization libraries can act as drop-in replacement for existing Dependency Injection containers, making PbO immediately applicable to a large number of enterprise software projects. The development of \textsc{ContainAnt} in this direction is a promising target of future work.

%%%%%%%%%%%%%%%%%%%%%%%%%%%%%
\bibliographystyle{plain}
\bibliography{dependency-injection-programming-arxiv}

\balance

\end{document}